\documentclass[lettersize,journal]{IEEEtran}
\usepackage{amsmath,amsfonts}
\usepackage{algorithmic}
\usepackage{array}
\usepackage[caption=false,font=normalsize,labelfont=sf,textfont=sf]{subfig}
\usepackage{textcomp}
\usepackage{stfloats}
\usepackage{url}
\usepackage{verbatim}
\usepackage{graphicx}
\usepackage{balance}
\usepackage{makecell}
\usepackage{color, colortbl}
\usepackage{multirow}
\usepackage{multicol}
\usepackage{booktabs}
\usepackage{hyperref}
\usepackage{cleveref}
\usepackage{cite}

\def\BibTeX{{\rm B\kern-.05em{\sc i\kern-.025em b}\kern-.08em
    T\kern-.1667em\lower.7ex\hbox{E}\kern-.125emX}}

\definecolor{Gray}{gray}{0.95}
\def\ie{\emph{i.e.}}

\newcommand{\modelname}{{MMFuser}}

\begin{document}
\title{MMFuser: Multimodal Multi-Layer Feature Fuser for Fine-Grained Vision-Language Understanding}
\author{Yue Cao,~Yangzhou Liu,~Zhe Chen,~Guangchen Shi,~Wenhai Wang,~Danhuai Zhao,~and Tong Lu \\

\IEEEcompsocitemizethanks{
This work was supported in part by the China Mobile Zijin Innovation Insititute under Grant NR2310J7M, in part by the National Natural Science Foundation of China under Grant 62372223, and in part by the Youth Student Basic Research Project of the National Natural Science Foundation for PhD Students under Grant 623B2050. \textit{(Yue Cao and Yangzhou Liu are co-first authors.) (Corresponding author: Tong Lu.)}

\IEEEcompsocthanksitem Yue Cao, Yangzhou Liu, Zhe Chen, Guangchen Shi, and Tong Lu are with the State Key Laboratory for Novel Software Technology, Nanjing University, Nanjing {\rm 210023}, China, (e-mail: caoyue0119@gmail.com; lyzlll2343@gmail.com; czcz94cz@gmail.com; guangchenshi@smail.nju.edu.cn; lutong@nju.edu.cn).
\IEEEcompsocthanksitem Wenhai Wang is with the Chinese University of Hong Kong, Hong Kong {\rm 999077}, China, (e-mail: wangwenhai362@163.com).
\IEEEcompsocthanksitem Danhuai Zhao is with the China Mobile Zijin Innovation Insititute, Nanjing {\rm 211899}, China, (e-mail: zhaodanhuai@chinamobile.com).
} 
}


\maketitle

\begin{abstract}
Despite significant advancements in Multimodal Large Language Models (MLLMs) for understanding complex human intentions through cross-modal interactions, capturing intricate image details remains challenging. 
Previous methods integrating multiple vision encoders to enhance visual detail introduce redundancy and computational overhead. 
We observe that most MLLMs utilize only the last-layer feature map of the vision encoder for visual representation, neglecting the rich fine-grained information in shallow feature maps. 
To address this issue, we propose \modelname, a simple yet effective multi-layer feature fuser that efficiently integrates deep and shallow features from Vision Transformers (ViTs). 
Specifically, it leverages semantically aligned deep features as queries to dynamically extract missing details from shallow features, thus preserving semantic alignment while enriching the representation with fine-grained information. 
Applied to the LLaVA-1.5 model, \modelname~achieves significant improvements in visual representation and benchmark performance, providing a more flexible and lightweight solution compared to multi-encoder ensemble methods. 
The code and model have been released at \url{https://github.com/yuecao0119/MMFuser}.

\end{abstract}


\section{Introduction}

\IEEEPARstart{I}{n} recent years, Multimodal Large Language Models (MLLMs)~\cite{instructblip,liu2023llava,liu2023llava_1_5,chen2023internvl,bai2023qwenvl,chen2024far,gpt4v} have emerged as a research hotspot in the field of Artificial General Intelligence (AGI). These models have made significant strides in understanding and expressing complex human intent through cross-modal interaction and learning. 
Building on rapid advancements in Large Language Models (LLMs)~\cite{chiang2023vicuna,touvron2023llama,touvron2023llama2,chatgpt,openai2023gpt4,dubey2024llama3,2023internlm,lu2024deepseek,qwen}, MLLMs utilize pre-trained vision encoders to extract image features and integrate them with advanced LLMs, demonstrating remarkable capabilities across a wide range of vision-language tasks.

\begin{figure}[t!]
    \centering
    \includegraphics[width=0.48\textwidth]{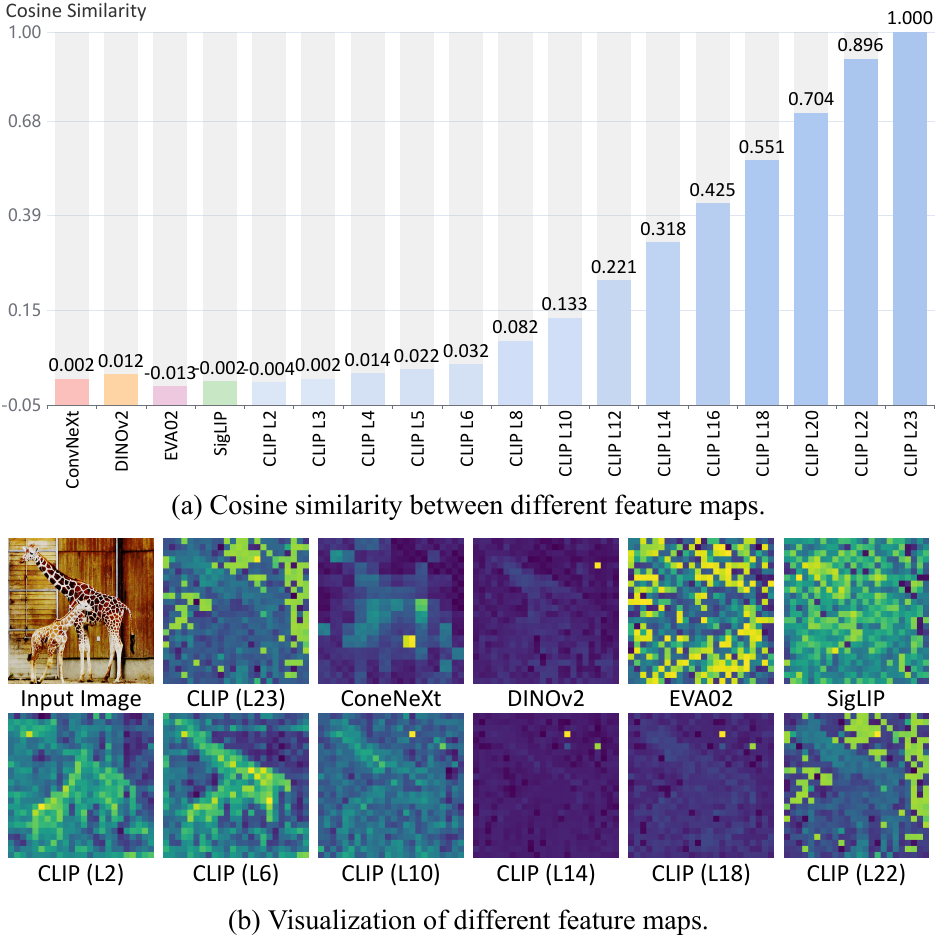}
    \caption{Comparison of feature maps from different vision encoders and various layers of CLIP-ViT. 
    (a) Cosine similarity is computed between the feature maps from various vision encoders, including CLIP-ViT-L~\cite{radford2021clip}, ConvNeXt-XXL~\cite{liu2022convnet}, DINOv2-L~\cite{oquab2023dinov2}, EVA02-L~\cite{fang2023eva02}, and SigLIP-L~\cite{zhai2023sigmoid}, and the final-layer feature map of CLIP-ViT-L.
    (b) Visualization of different feature maps.
    These results indicate significant feature differences not only between various vision encoders but also across different layers within the same vision encoder. This observation motivates us to fully explore the potential of individual vision encoders for developing MLLMs.
    }
    \label{fig:feature-visual-clip-layer}
\end{figure}

Currently, the mainstream approach~\cite{liu2023llava,liu2023llava_1_5,chen2023internvl,bai2023qwenvl,zheng2023minigpt5,zhu2023minigpt4,xu2024llavauhd} in the community involves using a pre-trained Vision Transformer (ViT)~\cite{dosovitskiy2020image,radford2021clip} or its variants~\cite{oquab2023dinov2,liu2022convnet,ge2024convllava} as the vision encoder, feeding the outputs from its final or penultimate layer into LLMs as visual representations. In this manner, these features with rich high-level semantic information are effectively transformed from an image space to a semantic text space. 
However, due to the loss of low-level image information in deep features, current MLLMs encounter challenges in accurately interpreting details, resulting in issues such as Optical Character Recognition (OCR) errors and object hallucinations.

To address these issues, recent studies~\cite{tong2024eyes,xu2024llavauhd} indicate that the ability of vision encoders to learn fine-grained image features has become a bottleneck for MLLMs. Consequently, some researchers~\cite{luo2024feast,jiang2024clipdinovisualencoders,li2024minigemini,lu2024deepseek} believe that solely relying on features of a single vision encoder may not be optimal. 
They propose integrating multiple pre-trained vision encoders, such as CLIP~\cite{radford2021clip}, DINOv2~\cite{oquab2023dinov2}, and ConvNext~\cite{liu2022convnet}, to enhance fine-grained visual representations in a complementary manner.
While these ensemble-based methods achieve promising results, they unfortunately introduce model redundancy and increase computational overhead.
Therefore, the necessity of employing multiple vision encoders remains contentious.

As a matter of fact, even with a single vision encoder, learned visual representations are diverse, as shown in Fig.~\ref{fig:feature-visual-clip-layer}.
The visualizations indicate that deep features are effective at extracting high-level semantic information, whereas shallow features are better suited for capturing low-level details such as edges and textures, which have not been fully leveraged in current MLLMs.
Looking back at classic image and video tasks such as object detection and semantic segmentation, multi-layer features are widely used~\cite{lin2017feature, 9868052, 9439490, 10154030,  wang2019fast}, where the combination of shallow and deep visual features provides a more comprehensive understanding of images or videos.
However, the application of multi-layer features in this manner remains relatively uncommon within the field of MLLMs.

It is natural to extend this idea to the visual representation of MLLMs. We have experimented with some straightforward methods to combine multiple layers of features from a single ViT~\cite{radford2021clip} for enhancing image detail, such as element-wise averaging or channel-wise concatenation.
However, these simple fusion methods have a little improvement in performance.
Through further exploration, we find that while shallow features capture richer low-level details, their semantic alignment with the text feature space is insufficient. 
This deficiency impairs the model's ability to understand images and offsets the benefits introduced by combining shallow and deep visual features, indicating that multi-layer feature fusion in MLLMs is vital and requires more thoughtful designs.

\begin{figure*}[t!]
    \centering
    \includegraphics[width=0.92\textwidth]{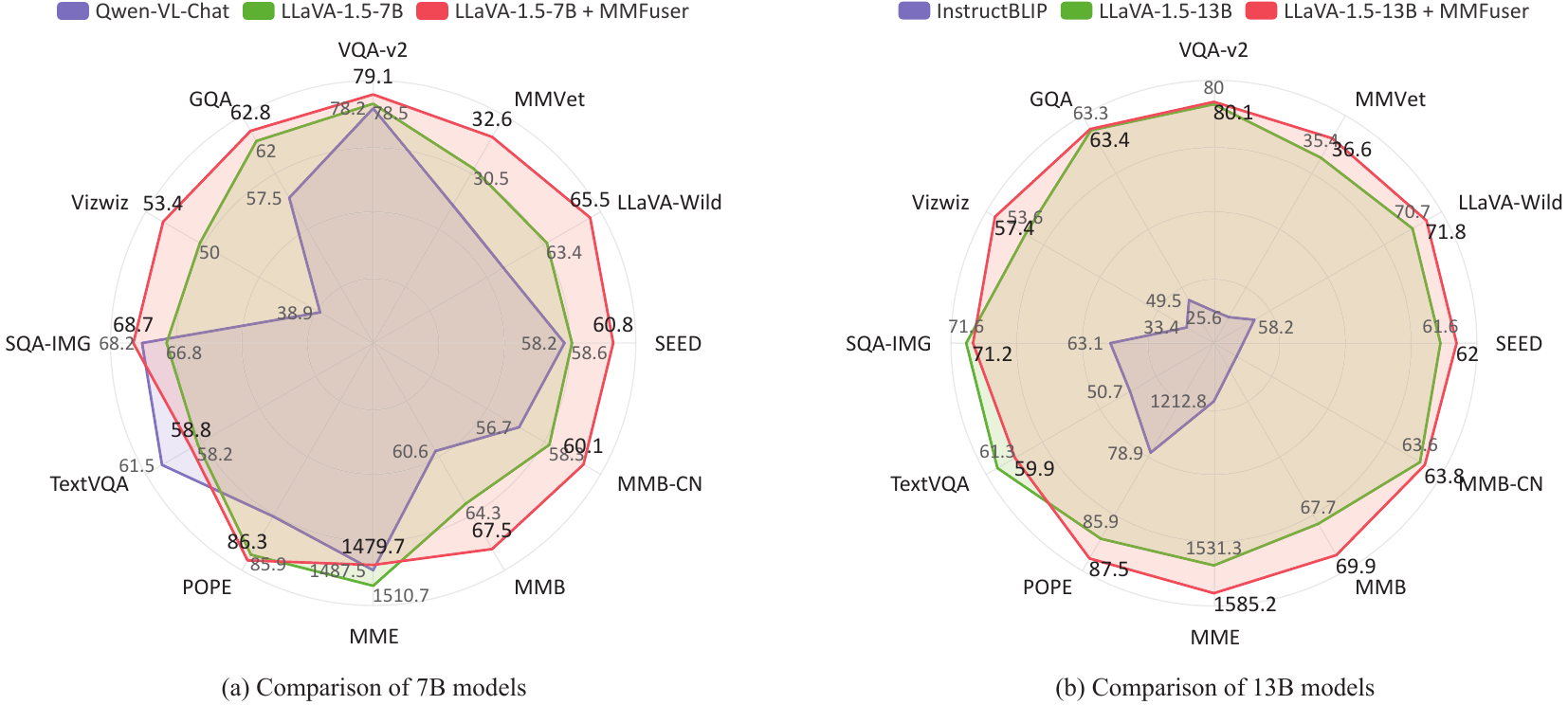}
    \caption{Performance comparison across different model sizes.
    (a) Among 7B models, including Qwen-VL-Chat~\cite{bai2023qwenvl}, LLaVA-1.5-7B~\cite{liu2023llava_1_5}, our model surpasses LLaVA-1.5-7B on 11 out of 12 benchmarks, with an average score of 61.8 compared to LLaVA-1.5-7B's 60.3. 
    (b) Among 13B models, including InstructBLIP~\cite{instructblip} and LLaVA-1.5-13B~\cite{liu2023llava_1_5}, our model also outperforms LLaVA-1.5-13B on 10 out of 12 benchmarks, achieving an average score of 64.1 compared to LLaVA-1.5-13B's 63.2.
    These results indicate that MMFuser can effectively improve the performance of LLaVA-1.5 models.
    }
    \label{fig:benchmark-compare}
\end{figure*}

Further, previous work~\cite{li2023blip2} suggests that LLMs excel in understanding deep features of ViTs, which are fully aligned with text feature spaces. Conversely, while shallow features are rich in details, they exhibit poor semantic alignment, making it challenging for LLMs to effectively interpret these features.
This insight inspired us to propose a simple yet effective method called \textbf{MMFuser} (see Fig.~\ref{fig:encoder-arch}), which uses deep features as queries to dynamically extract missing details from shallow features. It minimizes the risk of shallow features disrupting semantic alignment, maintaining the coherence of deep features while enriching them with fine-grained information. By leveraging multi-layer features, MMFuser can enhance the overall performance of MLLM in processing images and videos.

To validate the effectiveness of \modelname, we applied it to the recent well-known model, LLaVA-1.5~\cite{liu2023llava_1_5}. 
As shown in Fig.~\ref{fig:benchmark-compare}, our MMFuser significantly enhances the visual representations input into the MLLM, thereby improving LLaVA-1.5's performance on most multimodal benchmarks. Specifically, our 7B model outperformed LLaVA-1.5-7B in 10 out of 12 benchmarks, and our 13B model outperformed LLaVA-1.5-13B in 10 out of 12 benchmarks. Besides, our model demonstrated superior performance in fine-grained recognition tasks, including OCR and visual grounding.

In summary, our main contributions are as follows:

\begin{itemize}
    \item We reveal that the expressive potential of single vision encoders in MLLMs is underutilized. Shallow features, rich in detail, suffer from poor semantic alignment with text features, indicating that simple fusion methods are inadequate and require more advanced design.
    \item We introduce \modelname, which enhances the visual representations of a single vision encoder by dynamically integrating fine-grained details from shallow features while maintaining the semantic coherence of deep features.
    \item Applying MMFuser to LLaVA-1.5 models, we achieve significant performance improvements. Our 13B model surpasses LLaVA-1.5 by 3.8, 53.9, and 2.2 points on the VizWiz, MME, and MMBench-EN, respectively, demonstrating the efficacy of our method.
\end{itemize}

\section{Related Work}

\subsection{Multimodal Large Language Model} 

Multimodal Large Language Models (MLLMs) integrate visual representations from images with linguistic embeddings from text, thereby enhancing the models' capabilities in comprehending and generating language descriptions of visual content. Most open-source MLLMs employ architectures that include a pre-trained vision encoder, an LLM, and a cross-modal connector.
Early models, such as the BLIP series~\cite{li2023blip2,instructblip}, utilized the Q-Former module to align text and images, thus improving multimodal capabilities. Flamingo~\cite{alayrac2022flamingo} employed a gated cross-attention mechanism to integrate images and text. LLaVA-1.5~\cite{liu2023llava_1_5} adopted an MLP projector to connect the pre-trained vision encoder with the LLM.
InternVL~\cite{chen2023internvl, chen2024far} employed a dynamic resolution strategy, segmenting images into tiles and encoding both the tiles and the thumbnail view together. It then uses a pixel shuffle operation to reduce the number of visual tokens before integrating these features with the LLM through an MLP projector.

Additionally, private MLLMs such as the Gemini series~\cite{team2023gemini, reid2024gemini_1_5}, GPT-4V~\cite{gpt4v}, and Claude-3V series~\cite{claude3series2024}, along with open-source MLLMs like MiniGPT-4~\cite{zhu2023minigpt4}, Qwen-VL~\cite{bai2023qwenvl}, CogVLM~\cite{wang2023cogvlm}, the VisionLLM series~\cite{wang2023visionllm, wu2024visionllmv2}, and the All-Seeing series~\cite{wang2023all, wang2024allseeingv2}, among others~\cite{peng2023kosmos2}, have demonstrated robust multimodal capabilities. These models exemplify the powerful ability of MLLMs to understand, generalize, and reason with multimodal information, consistently setting new benchmarks in multimodal tasks.

\begin{figure}[t!]
    \centering
    \includegraphics[width=0.48\textwidth]{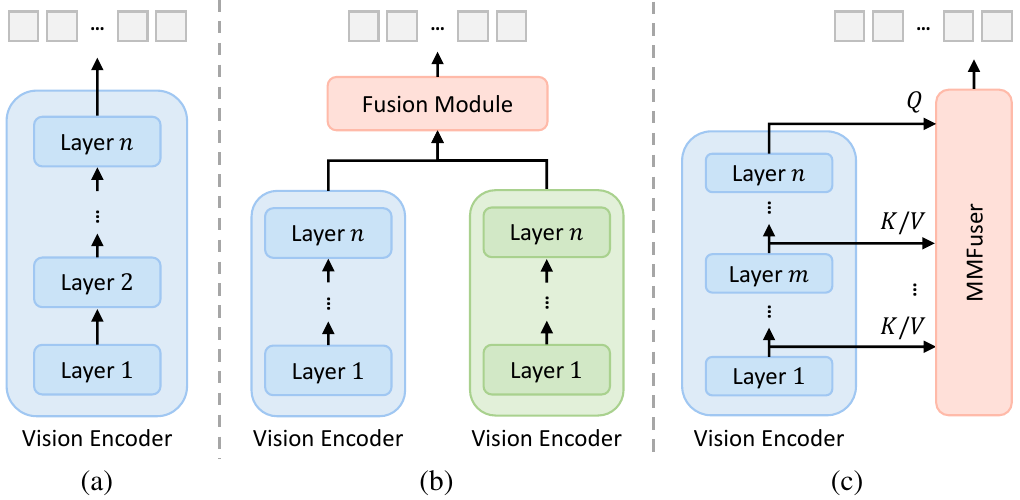}
    \caption{{Previous methods \emph{vs.} the proposed \modelname.} 
    (a) Previous methods typically utilize visual features from the final or penultimate layer of the vision encoder. For example, the LLaVA series~\cite{liu2023llava, liu2023llava_1_5} adopted this approach.
    (b) Some models integrate visual features from multiple vision encoders, such as MouSi~\cite{fan2024mousi}, DeepSeek-VL~\cite{lu2024deepseek}, and LLaVA-HR~\cite{luo2024feast}.
    (c) Our \modelname~fuses visual features from different layers of a single vision encoder, providing richer detail and better semantic alignment with text.
    }
    \label{fig:encoder-arch}
\end{figure}

\subsection{Vision Encoder in MLLMs}
The vision encoder plays a pivotal role in MLLMs, where notable models like CLIP-ViT~\cite{radford2021clip} have been widely used in this field. 
CLIP~\cite{radford2021clip} leveraged contrastive learning on large-scale image-text pairs for pre-training, resulting in a vision encoder that learns rich and general visual representations. This capability enhances the understanding of the relationship between vision and language.
Several models, including the LLaVA series~\cite{liu2023llava, liu2023llava_1_5}, PaLI~\cite{chen2022pali}, and Qwen-VL~\cite{bai2023qwenvl}, adopted CLIP-ViT~\cite{radford2021clip, openclip} as their default vision encoder.

Additionally, other vision foundation models are employed to construct MLLMs. For instance, CogVLM~\cite{wang2023cogvlm} leveraged the pre-trained EVA2-CLIP-E~\cite{sun2023evaclip} model for visual representations. ConvLLaVA~\cite{ge2024convllava} incorporated ConvNeXt~\cite{liu2022convnet}, a convolution-based hierarchical model, as its vision encoder. In DeepSeek-VL~\cite{lu2024deepseek}, SigLIP~\cite{zhai2023sigmoid} and SAM-ViT~\cite{kirillov2023sam} were utilized as vision encoders. Furthermore, InternVL~\cite{chen2023internvl, chen2024far} employed InternViT-6B, a vision foundation model trained on web-scale image-text data.
These works typically use the feature map from the final layer of the vision encoder as the visual representation, as shown in Fig.~\ref{fig:encoder-arch} (a). 
In contrast, our approach aims to explore the potential benefits of using feature maps from the shallow and intermediate layers of the vision encoder for vision-language tasks.

\subsection{Enhanced Visual Representation in MLLMs}
Many works are dedicated to enhancing visual representations in MLLMs, including:
\subsubsection{Scaling Up the Vision Encoder} 
PaLI~\cite{chen2022pali} increased the parameters of its vision encoder to 4 billion. In PaLI-17B, the vision encoder, ViT-e, accounts for approximately 25\% of the total parameters. 
InternVL~\cite{chen2023internvl} scaled its vision foundation model to 6 billion parameters, progressively aligning it with a large language model.
PaLM-E~\cite{driess2023palm} achieved a scale of 562 billion parameters by integrating the 540 billion parameter PaLM~\cite{chowdhery2023palm} LLM with the ViT-22B~\cite{dehghani2023scaling}.

\subsubsection{Integrating Multiple Vision Encoders}
As shown in Fig.~\ref{fig:encoder-arch} (b), this method enhances visual representations by integrating multiple vision encoders.
For example, MMVP~\cite{tong2024eyes} employed a Mixture of Features (MoF) approach to integrate image features from CLIP-ViT and DINOv2~\cite{oquab2023dinov2}. Notably, DINOv2 was a self-supervised vision model trained without any language guidance.
Similarly, MouSi~\cite{fan2024mousi} utilized an ensemble technique to synergize the capabilities of individual vision encoders. This method introduced a fusion network to unify the processing of outputs from different vision encoders, including CLIP, DINOv2, and SAM~\cite{kirillov2023sam}.
LLaVA-HR~\cite{luo2024feast} integrated image features from two vision encoders of different resolutions: 336px from CLIP-ViT and 1024px from CLIP-ConvNeXt~\cite{liu2022convnet}. This approach leveraged the strengths of both resolution inputs to enhance visual understanding.
DeepSeek-VL~\cite{lu2024deepseek} employed a hybrid vision encoder design, effectively encoding images by combining SigLIP-L~\cite{zhai2023sigmoid} for low-resolution inputs and SAM-B~\cite{kirillov2023sam} for high-resolution inputs.

\subsubsection{Feature Fusion}
MEP3P~\cite{10599309} enhanced the original visual features input into MLLMs with image depth features and pseudo-3D positions.
VCMR~\cite{10179965} utilized deformable attention to process multi-granularity image features and obtain fine-grained information, thereby improving performance in subsequent cross-modal tasks. 
Our concurrent work, Dense Connector~\cite{yao2024dense}, integrated features from multiple layers, enriching the visual inputs for MLLMs by capturing multi-level representations from the vision encoder.

Overall, these methods demonstrated significant performance boosts for MLLMs. However, the potential of a single vision encoder remained underexplored. To address this, we proposed MMFuser, to integrate feature maps from multiple layers of the vision encoder, as shown in Fig.~\ref{fig:encoder-arch} (c). This approach allowed us to obtain more powerful visual representations, thereby enhancing the performance of MLLMs.

\begin{figure*}[t!]
    \centering
    \includegraphics[width=\textwidth]{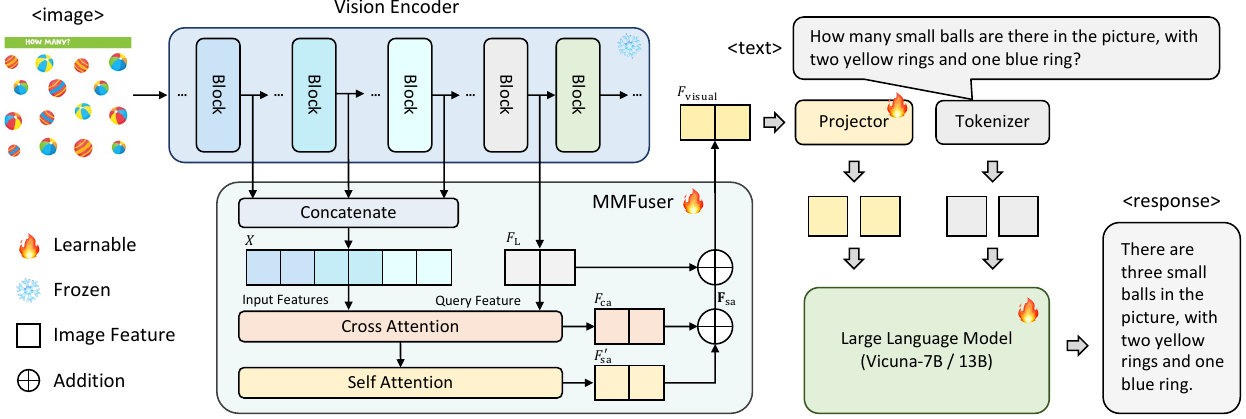}
    \caption{Overview of \modelname. In \modelname, feature maps from different layers of the vision encoder are strategically integrated to enhance the visual representations. Deep feature maps are employed as query elements, while shallow and intermediate feature maps are concatenated to form key and value elements. 
    Through a dynamic attention-based fusion, \modelname~combines fine-grained details and higher-level semantic information. The fused features are then aligned with text using a projector and subsequently passed as inputs to LLMs.} 
    \label{fig:model-arch}
\end{figure*}

\section{Method}

In this section, we address the challenge of missing detailed information in current visual representations for MLLMs. 
Initial attempts using some simple fusion methods demonstrate that shallow features suffer from inadequate semantic alignment. To overcome this, we propose the \modelname~for MLLMs, designed to effectively integrate multi-layer features while maintaining semantic alignment.

\subsection{Analysis of Visual Representations for MLLMs}

Currently, most mainstream MLLMs~\cite{liu2023llava,liu2023llava_1_5, chen2023internvl, zhu2023minigpt4, mckinzie2024mm1, li2024monkey} adopt CLIP-ViT~\cite{radford2021clip} as their vision encoder, typically selecting a single feature map from the final layers as the visual representation. 
Prior studies~\cite{raghu2021vision} suggest that in deeper layers of ViT, the receptive fields of attention heads become predominantly global, while the shallow layers retain both local and global information. Consequently, the lack of local details in deep feature maps can lead to suboptimal performance in fine-grained visual recognition tasks.

Different from existing approaches that integrate multiple encoders~\cite{luo2024feast,jiang2024clipdinovisualencoders,li2024minigemini,lu2024deepseek}, we consider that the visual information captured by the CLIP-ViT itself is not fully leveraged in MLLMs.
As illustrated in Fig.~\ref{fig:feature-visual-clip-layer}, shallow layers capture fine-grained details, which are often underutilized. 
Therefore, we argue that combining shallow and deep features can significantly enhance MLLM performance.

To validate our point, we explored several feature map fusion methods. We selected $L$ feature maps from various depths of ViT, denoted as $\mathcal{F} = [{F}_1, {F}_2,... ,{F}_{\rm L}]$, where ${F}_i \in \mathbb{R}^{N\times D}$, with $i \in [1, L]$. Here, $N$ represents the number of ViT patches, and $D$ denotes the dimension of the image feature embeddings. The four fusion methods considered are: \textit{(1) Concatenation}: Concatenate feature maps along the channel dimension to create the fused feature map, \ie~$\operatorname{Concat}({F}_1, {F}_2,... ,{F}_{\rm L})$. \textit{(2) Average}: Compute the element-wise average of all feature maps to obtain the fused feature map, \ie~$\frac{1}{L} \sum_{i=1}^{L} {F}_i$. \textit{(3) Weighted Average}: Assign learnable weights to each feature map, then compute the weighted average as $\sum_{i=1}^{L} w_i {F}_i$, where $w_i$ is the learnable weight associated with ${F}_i$. \textit{(4) Feature Pyramid Network~(FPN)}~\cite{lin2017feature}: Feed all feature maps into FPN for multi-scale feature learning, then compute the weighted average of the FPN outputs to obtain the fused feature map.

\begin{table}[t!]
    \centering
    \setlength\tabcolsep{0.95mm}
    \caption{Comparison between the LLaVA-1.5 Baseline, Four Simple Feature Fusion Methods, and Our MMFuser.}
    \label{tab:sample-fusion}
    \begin{tabular}{l | c c c c c | l}
    \toprule
    \multirow{2}{*}{Method}  & VizWiz & POPE & MME & MMB$^{\rm CN}$ & MMVet & \multirow{2}{*}{Avg.}  \\
           & \cite{gurari2018vizwiz} & \cite{li2023pope} & \cite{fu2023mme} & \cite{liu2023mmbench} & \cite{yu2023mmvet} \\
    \midrule
    LLaVA-1.5-13B~\cite{liu2023llava_1_5} & 53.6 & 85.9 & 1531.3 & 63.6 & 35.4 & 63.0 \\
    \midrule
    $w/$ {Concatenation}             &  52.1 & 86.9 & 1537.5 & 63.7 & 35.8 & 63.5 \\
    $w/$ {Average}                   &  54.7 & 87.1 & 1527.9 & 63.6 & 35.7 & 63.0 \\
    $w/$ {Weighted Average}          & 54.4 & 87.0 & 1532.8  & 62.5  & 34.6  & 63.1 \\
    $w/$ {FPN}~\cite{lin2017feature} & 53.7 & 87.3 & 1553.2  & 63.4  & 37.3  & 63.9\\
    \midrule
    $w/$ {MMFuser} (Ours) & 57.4 & 87.5 & 1585.2  & 63.8  & 36.6  & \textbf{64.9} \\
    \bottomrule
    \end{tabular}
\end{table}

We applied the fused feature maps as the visual representations in LLaVA-1.5~\cite{liu2023llava_1_5} and followed its original settings to evaluate model performance. 
However, as shown in Table~\ref{tab:sample-fusion}, none of the four fusion methods consistently improved model performance.
We attribute this to the semantic misalignment between deep and shallow features. 
As shown in Fig.~\ref{fig:feature-visual-clip-layer}, although shallow features capture more fine-grained details, their alignment with text is considerably weaker than that of deep features. 
In deeper layers, the features that correspond to the text are prominently highlighted, while shallow features lack this clear correspondence. 
By simply fusing the two types of image features, the model struggles to effectively leverage the complementary strengths of each feature.

\subsection{MMFuser: Multimodal Multi-Layer Feature Fuser}

Building on the insights from the previous sections, we observe that the shallow and deep features of ViT can complement each other. To harness this potential, we propose a multi-layer feature fusion module, \modelname. It can serve as a bridge between the vision encoder and the LLM. The overall architecture of \modelname~is shown in Fig.~\ref{fig:model-arch}.

Specifically, we extract $L$ feature maps from the ViT, denoted as $\mathcal{F} = [{F}_1, {F}_2,... ,{F}_{\rm L}]$. 
Since the strong semantic alignment between deep visual features and text space, we use the deep feature ${F}_{\rm L}$ as queries to dynamically extract missing details from shallow features ${X} = \operatorname{Concat}({F}_1, {F}_2,... ,{F}_{\rm L-1})$, through a cross-attention operation. This results in a visual feature ${F}_{\rm ca}\in\mathbb{R}^{N\times D}$ with richer fine-grained features. This process can be formulated as:
\begin{equation}
\begin{array}{c}
{F}_{\rm ca} = \operatorname{Attention}(\operatorname{norm}({F}_{\rm L}), \ \operatorname{norm}({X})),
\end{array}
\end{equation}
 where $\operatorname{Attention}(\cdot)$ denotes the attention mechanism, $\operatorname{norm}(\cdot)$ means layer normalization~\cite{ba2016layer}, and $\operatorname{Concat}(\cdot)$ represents the concatenation operation.

To effectively facilitate feature interaction and emphasize salient features, we incorporate a self-attention mechanism into the feature map ${F}_{\rm ca}$, formulated as:
\begin{gather}
{F}_{\rm sa}^{\prime} = \operatorname{Attention}(\operatorname{norm}{({F}_{\rm ca})},\ \operatorname{norm}{({F}_{\rm ca})}), \notag \\
{F}_{\rm sa} = {F}_{\rm ca} + \gamma_2 {F}_{\rm sa}^{\prime},
\end{gather}
where $\gamma_2 \in \mathbb{R}^{D}$ is a learnable vector that adjusts the contribution of ${F}_{\rm ca}$ relative to ${F}_{\rm sa}^{\prime}$. 
Subsequently, for the resulting feature map ${F}_{\rm sa}$, we introduce another learnable vector $\gamma_1 \in \mathbb{R}^{D}$ to modulate the integration of ${F}_{\rm L}$ and ${F}_{\rm sa}$:
\begin{equation}
\begin{array}{c}
{F}_{\rm visual} = {F}_{\rm L} + \gamma_1 {F}_{\rm sa}.
\end{array}
\end{equation}

Through the aforementioned steps, we derive the enhanced visual feature ${F}_{\rm visual}$. Unlike the original visual feature ${F}_{\rm L}$, ${F}_{\rm visual}$ integrates richer fine-grained information, making it a superior alternative for the visual input to the LLM.

\begin{table*}[t!]
    \setlength\tabcolsep{1.6mm}
    \centering
    \caption{{Comparison with state-of-the-art MLLMs on 12 general multimodal benchmarks.} 
    The benchmarks include: VQAv2~\cite{goyal2017making}, GQA~\cite{hudson2019gqa}, VizWiz~\cite{gurari2018vizwiz}, ScienceQA-IMG~\cite{lu2022sqa}; TextVQA~\cite{singh2019textvqa}, POPE~\cite{li2023pope}, MME~\cite{fu2023mme}, MMBench-EN~\cite{liu2023mmbench}, MMBench-CN~\cite{liu2023mmbench}, SEED-Bench~\cite{li2023seed}, LLaVA-Bench-In-the-Wild~\cite{liu2023llava}, and MMVet~\cite{yu2023mmvet}. $^*$~The training annotations of the datasets are observed during training. The best results are marked in \textbf{bold}, and the second best results are \underline{underlined}.}
    \label{tab:vqa_bench}
    \begin{tabular}{p{2.3cm}p{1.5cm} | ccccc | ccc c c c c }
    \toprule
    Method & LLM & VQA$^\text{v2}$ & GQA & VizWiz & SQA$^\text{I}$ & VQA$^\text{T}$ & POPE & MME & MMB & MMB$^\text{CN}$ & SEED & LLaVA$^\text{W}$ & MMVet \\
    \midrule
    InstructBLIP~\cite{instructblip} & Vicuna-7B   & -- & 49.2 & 34.5 & 60.5 & 50.1 & -- & --& 36.0 & 23.7 & 53.4& 60.9 & 26.2 \\
    IDEFICS-9B~\cite{idefics2023} & LLaMA-7B & 50.9 & 38.4 & 35.5 & -- & 25.9 & -- & -- & 48.2 & 25.2 & -- & -- & -- \\
    Qwen-VL~\cite{bai2023qwenvl} & Qwen-7B & ~\underline{78.8}$^*$ & ~{59.3$^*$} & 35.2 & 67.1 & ~\textbf{63.8}$^*$ & -- & -- & 38.2 & 7.4 & 56.3 & -- & -- \\
    Qwen-VL-Chat~\cite{bai2023qwenvl} & Qwen-7B & ~78.2$^*$ & ~57.5$^*$ & 38.9 & \underline{68.2} & ~\underline{61.5}$^*$ & -- & \underline{1487.5} & 60.6 & 56.7 & 58.2 & -- & -- \\   
    {LLaVA-1.5-7B}~\cite{liu2023llava_1_5} & Vicuna-7B & ~{78.5}$^*$ & ~\underline{62.0}$^*$ & \underline{50.0} & 66.8 & 58.2 & \underline{85.9} & \textbf{1510.7} & \underline{64.3} & \underline{58.3} & \underline{58.6} & \underline{63.4} & \underline{30.5} \\
    \rowcolor{Gray}
    \makecell[l]{{LLaVA-1.5-7B}\\ {+ {\modelname}~(Ours)}} & Vicuna-7B & ~\textbf{79.1}$^*$ & ~\textbf{62.8}$^*$ & \textbf{53.4} & \textbf{68.7} & 58.8 & \textbf{86.3} & {1479.7} & \textbf{67.5} & \textbf{60.1}  & \textbf{60.8} & \textbf{65.5} & \textbf{32.6} \\
    \midrule
    BLIP-2~\cite{li2023blip2} & Vicuna-13B & 65.0 & 41.0 & 19.6 & 61.0 & 42.5 & 85.3 & 1293.8 & --  & -- & 46.4 & 38.1 & 22.4 \\
    InstructBLIP~\cite{instructblip} & Vicuna-13B & -- & 49.5 & 33.4 & 63.1 & 50.7 & 78.9 & 1212.8 & -- & -- & -- & 58.2 & 25.6 \\
    IDEFICS-80B~\cite{idefics2023} & LLaMA-65B & 60.0 & 45.2 & 36.0 & -- & 30.9 & -- & -- & 54.5 & 38.1 & -- & -- & -- \\
    Shikra~\cite{chen2023shikra} & Vicuna-13B& ~77.4$^*$ & -- & -- & -- & -- & -- & -- & 58.8 & -- & -- & -- & -- \\
    {LLaVA-1.5-13B}~\cite{liu2023llava_1_5} & Vicuna-13B & ~\underline{80.0}$^*$ & ~\underline{63.3}$^*$ & \underline{53.6} & \textbf{71.6} & \textbf{61.3} &  \underline{85.9} & \underline{1531.3} & \underline{67.7} & \underline{63.6} & \underline{61.6} & \underline{70.7} & \underline{35.4} \\
    \rowcolor{Gray}
    \makecell[l]{{LLaVA-1.5-13B}\\ {+ {\modelname}~(Ours)}} & Vicuna-13B & ~\textbf{80.1}$^*$ & ~\textbf{63.4}$^*$ & \textbf{57.4} & \underline{71.2} & \underline{59.9} & \textbf{87.5} & \textbf{1585.2} & \textbf{69.9} & \textbf{63.8}  & \textbf{62.0} & \textbf{71.8} & \textbf{36.6} \\
    \bottomrule
    \end{tabular}
\end{table*}

\subsection{Overall Framework Design}

Our proposed \modelname~can be integrated into mainstream open-source MLLMs, particularly within the ``ViT-MLP-LLM'' architecture~\cite{liu2023llava, liu2023llava_1_5, chen2023internvl, zhu2023minigpt4}. 
As a case study, we demonstrate this integration using the LLaVA-1.5 model.
In this framework, \modelname~is positioned between the ViT and the MLP projector to fuse multi-layer feature maps from the ViT.
The overall architecture, illustrated in Fig.~\ref{fig:model-arch}, consists of four key components: the vision encoder (ViT), \modelname, MLP projector, and the large language model.

The input image is first fed into the ViT to extract multi-layer visual features. Then, our \modelname~leverages these multi-layer features to obtain a fused feature with richer local information. Following the pipeline in LLaVA-1.5~\cite{liu2023llava_1_5}, the fused features are then aligned with the text embedding space through a trainable MLP projector. 
Similarly, the input text is transformed into a text embedding via a tokenizer. Finally, the image and text embeddings are concatenated and fed into an LLM, such as Vicuna-7B~\cite{chiang2023vicuna}. The LLM then answers the user's questions based on the image features. This framework can also be readily adapted for processing video data. Overall, existing mainstream MLLMs can easily adopt our \modelname~to enhance their visual feature extraction capabilities.

\section{Experiment}

\subsection{Implementation Details}

We adopt LLaVA-1.5~\cite{liu2023llava_1_5} as the baseline to study the visual representations of MLLMs. 
The model comprises three components: the pre-trained vision encoder CLIP-ViT-L-336px~\cite{radford2021clip}, the pre-trained LLM Vicuna-v1.5~\cite{chiang2023vicuna}, and a two-layer MLP projector.
To fully leverage the potential of a single vision encoder, we use the proposed \modelname~to fuse multi-layer features from the vision encoder, replacing the original single-layer image feature for the LLM.

\subsubsection{Architecture Settings}
In \modelname, the number of feature layers $L$ selected from the ViT is set to 5 by default. The parameters $\gamma_1$ and $\gamma_2$, which control the weights of features from different layers, are both initialized to 0. Deformable attention~\cite{zhu2020deformable} is used as the default attention mechanism in \modelname. In this setup, the number of sampling points is fixed at 4, with the attention layer employing 16 heads.

\subsubsection{Training Settings}
For a fair comparison, we adopt the same two-stage training recipes as LLaVA-1.5~\cite{liu2023llava_1_5}:

\textbf{Pre-training.} 
During the pre-training stage, we utilize the LLaVA-LCS-558K dataset~\cite{liu2023llava}, which comprises 558K image-caption pairs. 
In this phase, the vision encoder and the LLM are kept frozen.
Training is focused solely on the MLP projector and \modelname, with the objective of aligning the visual features with the input space of the LLM.

\textbf{Fine-tuning.} 
During the fine-tuning stage, we employ the LLaVA-Instruct-665K dataset~\cite{liu2023llava_1_5}, which includes 665K instruction-following data from sources such as GQA~\cite{hudson2019gqa}, TextCaps~\cite{sidorov2020textcaps}, ShareGPT~\cite{sharegpt}, and others~\cite{goyal2017making, marino2019ok, mishra2019ocrvqa, schwenk2022aokvqa, kazemzadeh2014refcoco, krishna2017vg}. In this phase, the MLP projector, \modelname, and LLM are trained end-to-end, while the vision encoder remains frozen.

We train our model using the same experimental settings and hyperparameters as LLaVA-1.5. We adopt the AdamW optimizer and use a cosine decay learning rate scheduler with an initial warmup ratio of 0.03 and no weight decay. The global batch size is set to 256 for pre-training and 128 for fine-tuning. The learning rate is set to 1e-3 during the pre-training phase and reduced to 2e-5 during the fine-tuning phase. Both phases are trained for a single epoch.

\subsection{Results on General Multimodal Benchmarks} 

We evaluated \modelname~using a comprehensive suite of 12 benchmarks, including both academic visual question answering (VQA) benchmarks and comprehensive multimodal benchmarks, to assess its performance across multiple dimensions.
As shown in Table~\ref{tab:vqa_bench}, our model exhibits substantial performance improvements on these benchmarks.

\subsubsection{Results on Academic VQA Benchmarks}
On the academic VQA benchmarks, our 7B model consistently outperforms LLaVA-1.5-7B across all five benchmarks. Similarly, the 13B version of our model surpasses LLaVA-1.5-13B on the VQAv2, GQA, and VizWiz benchmarks, with a particularly notable improvement of 3.8 points on VizWiz. Furthermore, our model achieves comparable performance on the ScienceQA and TextVQA benchmarks.

\begin{table}[t!]
    \setlength\tabcolsep{1.35mm}
    \centering
    \caption{Result on OCRBench~\cite{liu2024hidden}. 
    Recog.: text recognition, VQA$^{\rm S}$: Scene Text-Centric VQA; VQA$^{\rm D}$: Document-Oriented VQA; KIE: Key Information Extraction; HMER: Handwritten Mathematical Expression; Final: Overall score across all five categories.
    }
    \label{tab:ocrbench}
    \begin{tabular}{l | ccccc | l}
    \toprule
     Method & {Recog.} & VQA$^{\rm S}$ & VQA$^{\rm D}$ & KIE  & HMER & Final   \\
    \midrule
    MiniGPT4V2~\cite{chen2023minigptv2} & 124 & 29 & 4 & 0 & 0 & 157 \\
     BLIP2~\cite{li2023blip2} & 154 & 71 & 10 & 0 & 0 & 235 \\
     InstructBLIP~\cite{instructblip} & 168 & 93 & 14 & 1 & 0 & 276 \\
    BLIVA~\cite{hu2024bliva} & 165 & 103 & 22 & 1 & 0 & 291 \\
    \midrule
    LLaVA-1.5-7B~\cite{liu2023llava_1_5} &  160 &	117 & 15 & 5 & 0 &  297  \\
    \rowcolor{Gray}
    + \modelname~(Ours) & 159 & 128 & 20 & 8 & 0 & \textbf{315}$^{\uparrow{18}}$ \\ 
    \midrule
    LLaVA-1.5-13B~\cite{liu2023llava_1_5} & {176} & 129 & 19 & 7 & 0 &  331 \\ 
    \rowcolor{Gray}
    + \modelname~(Ours) & 171	& {136} & {25} & {11}& 0 & \textbf{343}$^{\uparrow{12}}$ \\
    \bottomrule
    \end{tabular}
\end{table}

\subsubsection{Results on Comprehensive Multimodal Benchmarks}
In the multimodal benchmarks, our 7B and 13B models demonstrate significant performance improvements compared to the corresponding LLaVA-1.5 models. Notably, our 13B model substantially outperforms LLaVA-1.5-13B across seven different benchmarks. Specifically, our 13B model achieves scores of 1585.2 on the MME benchmark and 69.9 on the MMBench benchmark, representing enhancements of 53.9 and 2.2 points over LLaVA-1.5-13B, respectively. Additionally, our model exhibits strong performance across other multimodal benchmarks, such as POPE, SEED-Bench, and MMVet.

\subsection{Results on OCRBench}

OCRBench~\cite{liu2024hidden} serves as a comprehensive OCR benchmark, comprising 1,000 manually curated and corrected OCR-related VQA instructions. The benchmark is systematically divided into five distinct categories: Text Recognition (Recog.), Scene Text-Centric VQA (VQA$^{\rm S}$), Document-Oriented VQA (VQA$^{\rm D}$), Key Information Extraction (KIE), and Handwritten Mathematical Expression Recognition (HMER).

As detailed in Table~\ref{tab:ocrbench}, our models, with 7B and 13B parameters, exhibit an average improvement of 15 points over LLaVA-1.5. This substantial gain underscores \modelname's enhanced capability in refining the granularity of visual representations, thereby contributing to more accurate text recognition and superior OCR performance.

\begin{table}[t]
\setlength\tabcolsep{0.95mm} 
\footnotesize
\centering
\caption{Results of Region Captioning. 
Results are reported with the CIDEr score.
}
\label{tab:cider}
\begin{tabular}{l|cccc cccc|c}
\toprule
\multirow{2}{*}{Model} & \multicolumn{3}{c}{RefCOCO} & \multicolumn{3}{c}{RefCOCO+} & \multicolumn{2}{c|}{RefCOCOg} & \multirow{2}{*}{Avg.}\\
 & val & testA & testB & val & testA & testB & val& test &   \\
\midrule
LLaVA-1.5-7B & 30.4 & 16.0 & 42.0 & 30.2 & 20.3 & 39.1 & 60.5 & 58.9 & 37.2 \\
\rowcolor{Gray}
 + {\modelname} & 33.6 & 17.7 & 45.9 & 33.6 & 21.2 & 42.6 & 61.5 & 61.6 & \textbf{39.7} \\
\midrule
LLaVA-1.5-13B & 33.1 & 16.7 & 45.2 & 33.4 & 19.8 & 41.6 & 61.6 & 59.9 & 38.9 \\
\rowcolor{Gray}
 + {\modelname} & 38.2 & 19.5 & 53.6 & 36.8 & 22.8 & 43.8 & 64.5 & 63.4 & \textbf{42.8} \\
\bottomrule
\end{tabular}
\end{table}

\begin{table}[t]
\setlength\tabcolsep{0.95mm}
\footnotesize
\centering
\caption{Results of Referring Expression Comprehension.
Results are reported with the Precision@0.5 score.
}
\label{tab:rec}
\begin{tabular}{l|cccc cccc|c}
\toprule
\multirow{2}{*}{Model} & \multicolumn{3}{c}{RefCOCO} & \multicolumn{3}{c}{RefCOCO+} & \multicolumn{2}{c|}{RefCOCOg} & \multirow{2}{*}{Avg.}\\
 & val & testA & testB & val & testA & testB & val& test &   \\
\midrule
LLaVA-1.5-7B & 56.2 & 64.4 & 47.5 & 50.0 & 59.2 & 39.0 & 48.8 & 48.4 & 51.7  \\
\rowcolor{Gray}
 + {\modelname} & 62.0 & 70.7 & 52.4 & 55.6 & 65.1 & 44.2 & 54.0 & 54.8 & \textbf{57.4}  \\
\midrule
LLaVA-1.5-13B &  66.5 & 73.9 & 55.7 & 59.8 & 67.9 & 48.7 & 57.3 & 56.0 & 60.7 \\
\rowcolor{Gray}
 + {\modelname} &  66.6 & 73.9 & 56.3 & 61.3 & 68.8 & 49.2 & 56.5 & 56.5 & \textbf{61.1} \\
\bottomrule
\end{tabular}
\end{table}

\subsection{Results on Region-level Benchmarks}
To assess regional understanding and grounding capabilities, we evaluate \modelname~on two representative regional-level tasks. (1) Regional Captioning~\cite{krishna2017vg, mao2016refcoco_plus_g}: This task requires the model to generate a description for an object in the image based on a given region. (2) Referring Expression Comprehension~\cite{kazemzadeh2014refcoco, mao2016refcoco_plus_g}: This task requires the model to locate target objects in an image based on a given description.

\subsubsection{Results of Region Captioning}

On region captioning tasks, our model shows significant improvements. As shown in Table~\ref{tab:cider}, compared to LLaVA-1.5, the 7B model of \modelname~surpasses LLaVA-1.5 by 2.5 points on average, while the 13B version improves by 3.9 points. 
This indicates that \modelname~captures fine-grained information, enhancing caption accuracy and richness.

\subsubsection{Results of Referring Expression Comprehension (REC)} 
We also employ REC tasks to evaluate the model's grounding capabilities. As shown in Table~\ref{tab:rec}, our model consistently outperforms LLaVA-1.5 models across all benchmarks, with an especially notable average improvement of 5.7 points for the 7B model compared to LLaVA-1.5-7B. 
This highlights that the visual representations generated by \modelname~are more detailed and comprehensive, enhancing spatial localization and significantly boosting performance in grounding tasks.

\begin{figure*}[t!]
    \centering
    \includegraphics[width=0.95\textwidth]{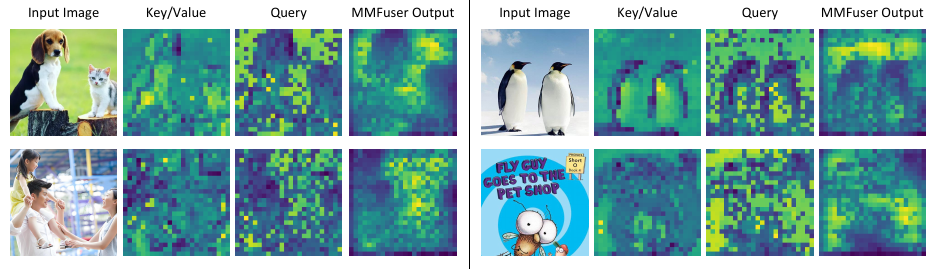}
    \caption{{Feature map visualization of \modelname.} 
    For each image, we provide three types of output feature maps.
    The term ``Key/Value'' refers to the averaged feature maps from four selected shallow and intermediate layers of the ViT—specifically, layers 3, 8, 13, and 18—used as the key and value inputs in MMFuser. 
    ``Query'' denotes the feature map from the penultimate layer of the ViT, serving as the query input in \modelname~and as the visual representations in prior MLLMs. ``MMFuser Output'' represents the feature map generated after applying the proposed MMFuser. 
    As can be seen, the proposed MMFuser captures fine-grained details from shallow and intermediate ViT layers, enriching the visual representations for the LLM.
    }
    \label{fig:mmfuser-visualization}
\end{figure*}

\subsection{Ablation Study}

\begin{table}[t]
\setlength\tabcolsep{0.8mm}
\centering
\caption{Ablation on layer combination.}
\label{tab:ablation_layer}
\begin{tabular}{l c|c c c c c c | l}
\toprule
\multicolumn{2}{l|}{Selected Layers} & \multirow{2}{*}{VizWiz} &\multirow{2}{*}{POPE} & \multirow{2}{*}{MME} & \multirow{2}{*}{MMB}  & \multirow{2}{*}{SEED} & \multirow{2}{*}{MMVet} & \multirow{2}{*}{Avg.} \\
$Q$ & \multicolumn{1}{c|}{$K,V$} &  &  &  &   &  & \\
\midrule
\multicolumn{1}{l}{--} & \multicolumn{1}{c|}{--} & 53.6 & 85.9 & 1531.3 & 67.7 & 61.6 & 35.4 & 63.5 \\
\midrule
23 & [1, 3, 5, 7] &  54.3 & 87.3 & 1582.2 & 69.5 & 61.8 & 35.1 & 64.5 \\
23 & [9, 11, 13, 15] & 52.4 & 87.0 & 1560.3 & 69.1 & 61.7 & 35.6 & 64.0 \\
23 & [17, 19, 21, 24] &  54.7 & 87.3 & \textbf{1591.2} & 69.3 & 62.0 & 35.8 & 64.8 \\
23 & [5, 8, 11, 20] & 54.7 & 87.2 & 1584.0 & 69.3 & \textbf{62.5} & 36.2 & 64.9  \\
\rowcolor{Gray}
23 & [3, 8, 13, 18] & \textbf{57.4} & \textbf{87.5} & {1585.2} & \textbf{69.9} & 62.0 & \textbf{36.6} & \textbf{65.4} \\
\bottomrule
\end{tabular}
\end{table}

\begin{table}[t]
\setlength\tabcolsep{0.2mm}
\centering
\caption{Ablations on attention mechanisms. ``Cplx.'' indicates the complexity type of the attention mechanism, encompassing both quadratic (Quad.) and linear variants.}
\label{tab:ablation_attn}
\begin{tabular}{l|c|cccccc | l}
\toprule
{Attention Type}  & {Cplx.}  & {VizWiz} &{POPE} & {MME} & {MMB}   & {SEED} & MMVet  & Avg.  \\
\midrule
\makecell[l]{LLaVA-1.5-13B~\cite{liu2023llava_1_5} } & -- & 53.6 & 85.9 & 1531.3 & 67.7 & 61.6 & 35.4 & 63.5 \\
\midrule
Global Attn~\cite{vaswani2017attention}  & Quad. & 52.9 & \textbf{87.6} & 1566.3 & 68.6 & \textbf{62.2} & 35.3  & 64.2 \\
\makecell[l]{Linear SRA~\cite{wang2021pvtv2}}  & Linear & 54.3 & 87.0 & 1581.6 & 68.9 & 61.9 & 34.7  &  64.3 \\
\rowcolor{Gray}
Deformable Attn~\cite{zhu2020deformable} & Linear & \textbf{57.4} & {87.5} & \textbf{1585.2}  & \textbf{69.9}  & 62.0 & \textbf{36.6} & \textbf{65.4} \\
\bottomrule
\end{tabular}
\end{table}

\begin{table}[t!]
\setlength\tabcolsep{1.1mm} 
\footnotesize
\centering
\caption{Ablations on internal module designs.}
\label{tab:ablation_module}
\begin{tabular}{l |c c c c c c | l}
\toprule
{Module} & {VizWiz} & {POPE} & {MME} & {MMB}  & {SEED} & {MMVet} & Avg.\\
\midrule
LLaVA-1.5-13B & 53.6 & 85.9 & 1531.3 & 67.7 & 61.6  & 35.4 & 63.5 \\
\midrule
+ Cross-Attn & 54.6 & 87.2 & 1557.1 & 68.2 & \textbf{62.2}  & 36.1 & 64.4 \\ 
\rowcolor{Gray}
\text{++} Self-Attn & \textbf{57.4} & \textbf{87.5} & \textbf{1585.2}  & \textbf{69.9}  & 62.0 & \textbf{36.6}  & \textbf{65.4}\\
\bottomrule
\end{tabular}
\end{table}

\subsubsection{Ablation Study on Layer Combination} 
To evaluate the contribution of feature maps from different layers of ViT~\cite{radford2021clip} to model performance, we conduct an ablation study on the \modelname-13B model by sampling feature maps from various depths. 
As detailed in Table~\ref{tab:ablation_layer}, we utilize the feature map from the 23rd layer of ViT as the query, while the key and value are formed by concatenating feature maps sampled from different layers: shallow (row 2), intermediate (row 3), deep (row 4), non-uniform (row 5), and uniform sampling (row 6). 

The results demonstrate that incorporating feature maps from multiple layers significantly improves the model’s visual representation capabilities. 
The feature maps from different layers capture information at varying receptive field sizes, each offering distinct advantages depending on the task. Notably, uniform sampling across layers, which aggregates information from diverse receptive fields, achieves the highest performance. Therefore, \modelname~employs this strategy to optimize visual representation for visual-language tasks.

\begin{figure*}[t!]
    \centering
    \includegraphics[width=\textwidth]{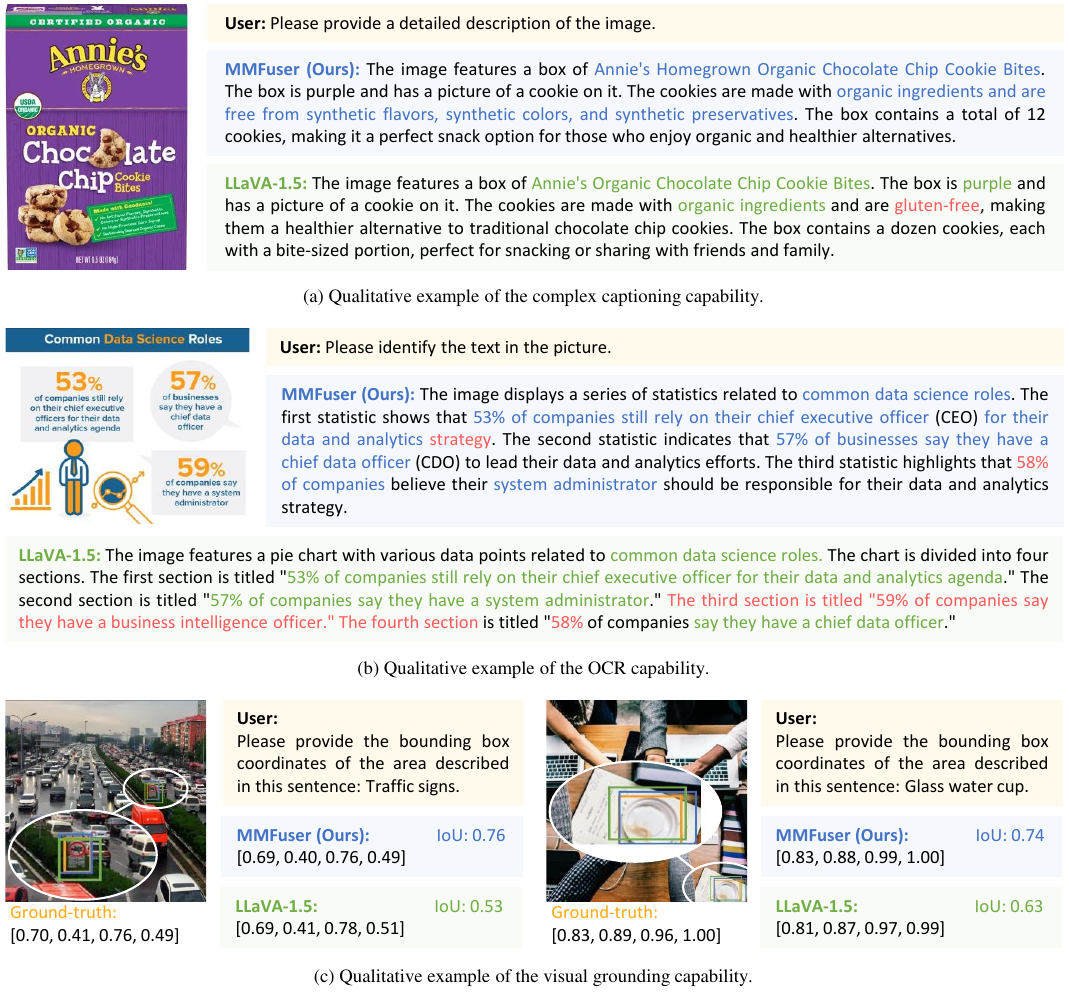}
    \caption{{Qualitative comparisons between \modelname~and LLaVA-1.5.} 
    For the complex captioning and OCR tasks, the text generated by each model is color-coded to match the model name, indicating content that accurately reflects the information present in the image. Text in red denotes errors or hallucinations. For visual grounding examples, the predicted bounding boxes are also color-coded according to the respective model names, while yellow bounding boxes denote the ground truth. The Intersection over Union (IoU) metric is used to evaluate the overlap between predicted bounding boxes and the ground truth. A higher IoU value indicates a more accurate prediction, as it reflects a larger intersection area relative to the union area.
    }
    \label{fig:visualization-chat}
\end{figure*}

\subsubsection{Ablation on Attention Mechanisms}
The attention mechanism in \modelname~is modular and can be replaced with different variants. We experiment with three types of attention mechanisms using the 13B model of \modelname. As shown in Table~\ref{tab:ablation_attn}, our framework consistently enhances the visual representation capability of the MLLM, regardless of the attention mechanism employed. This confirms the effectiveness and adaptability of our feature fusion module.

Compared to global attention mechanisms~\cite{vaswani2017attention}, sparse attention mechanisms with linear complexity not only deliver superior performance in \modelname~but also offer substantial improvements in computational efficiency. 
Among the linear attention mechanisms, deformable attention~\cite{zhu2020deformable} provides the most significant performance gains. Therefore, deformable attention is adopted as the default mechanism in \modelname. It is noteworthy that future research could explore more advanced attention mechanisms to further enhance the visual representation capabilities of MLLMs.

\subsubsection{Ablation on Internal Module Designs} 
To validate the key contributions of our \modelname~architecture, we incrementally enhance the LLaVA-1.5-13B baseline~\cite{liu2023llava_1_5} with our proposed designs. As illustrated in Table~\ref{tab:ablation_module}, integrating cross-attention mechanisms to extract fine-grained information from various ViT layers leads to a marked performance improvement. Specifically, our model outperforms the baseline by 1.3 points on POPE and 0.7 points on MMVet. Moreover, the addition of self-attention to these fine-grained features further enhances the model's capability to capture relevant information, resulting in gains of 2.8 points on VizWiz, 28.1 points on MME, and 1.7 points on MMB. These results collectively demonstrate that our design significantly enhances the visual representation abilities of MLLMs, highlighting the critical role of each component in our model.

\subsubsection{Visual Representation Visualization}
To intuitively validate the impact of \modelname~on visual features, we present the input and output feature map visualizations for four example images in Fig.~\ref{fig:mmfuser-visualization}. 
For the input image, the shallow feature maps convey richer fine-grained information, but this information is messy and the semantic information is difficult to distinguish. For instance, the lower-left image contains complex information, making it challenging to intuitively discern semantic content from the shallow features. But after the attention mechanism of \modelname, the foreground characters in the picture are well highlighted, and the semantic information is aligned with the last layer feature map of ViT. The aligned detailed features can effectively enhance the fine-grained perception capability of MLLMs.

\subsection{Qualitative Comparison}

In Fig.~\ref{fig:visualization-chat}, we compare \modelname~and LLaVA-1.5 across three case studies related to fine-grained analysis, including complex captioning, OCR, and visual grounding. The visualization results show that the rich fine-grained information in the shallow feature maps significantly enhances the ability of MLLMs to capture and process image details. 

For the complex captioning example, \modelname~shows great improvement in recognizing small text and small objects. Compared to LLaVA-1.5, \modelname~even successfully identifies the small and blurred green background text describing the healthy ingredients of the cookie: ``free from synthetic flavors, synthetic colors, and synthetic preservatives."

For the OCR task, \modelname~shows enhanced recognition of small texts, effectively mitigating hallucinations that often occur in LLaVA-1.5. The latter struggles with misaligned and hallucinated content due to inadequate text recognition capabilities, leading it to supplement details from its prior knowledge rather than the image itself.

In the visual grounding example, \ demonstrates a more accurate delineation of object boundaries, especially when adjacent objects have similar colors. This improvement is more pronounced with small objects, such as the positioning of the traffic sign in the left-side example, where \modelname~shows a 23\% increase in the IoU metric.

In summary, these results further validate the effectiveness of \modelname~in addressing the issue of missing fine-grained image details in MLLMs.

\section{Conclusion}

In this paper, we introduced \modelname, a novel multimodal multi-layer feature fuser designed to improve visual representation in Multimodal Large Language Models (MLLMs). By integrating both shallow and deep features from the vision encoder, \modelname~addresses the limitations of relying solely on deep features from a single vision encoder, which often results in the loss of fine-grained details. Our experiments show that \modelname~enhances the performance of the LLaVA-1.5 model across various benchmarks, enriching visual representation without the redundancy and computational overhead of ensemble models. This approach maximizes the potential of a single ViT encoder, offering an efficient and flexible solution for MLLMs. Overall, \modelname~improves fine-grained detail capture and semantic understanding in MLLMs, and we hope it will contribute to the community's efforts in developing more robust and efficient multimodal models.

\bibliographystyle{IEEEtran}
\bibliography{egbib}

\end{document}